\documentclass[sigconf]{acmart}

\usepackage{booktabs} 
\usepackage{graphicx}
\usepackage{color}
\usepackage{colortbl}
\usepackage{boldline}
\usepackage{hhline}
\usepackage{arydshln}
\usepackage{mathtools}
\usepackage{graphicx}
\usepackage{color}
\usepackage{ifthen}
\usepackage{multirow}
\usepackage{arydshln}

\makeatletter
\@namedef{ver@everyshi.sty}{}
\makeatother
\usepackage{tikz}
\usepackage{subfigure}

\editor{Anoop M. Namboodiri}
\editor{Vineeth Balasubramanian}
\editor{Amit Roy-Chowdhury}
\editor{Guido Gerig}

\newcommand*{\origrightarrow}{}

\renewcommand*{\textrightarrow}{\fontfamily{cmr}\selectfont\origrightarrow}

\begin{document}
\title{Epipolar Geometry based Learning of Multi-view Depth and Ego-Motion from Monocular Sequences}

\author{Vignesh Prasad, Dipanjan Das, Brojeshwar Bhowmick}
\affiliation{%
  \institution{Embedded Systems \& Robotics, TCS Research \& Innovation, Kolkata}
  \streetaddress{}
  \city{}
  \state{}
  \postcode{}
}
\email{{vignesh.prasad, dipanjan.da, b.bhowmick}@tcs.com}

\renewcommand{\shortauthors}{Prasad et al.}
\renewcommand{\shorttitle}{Epipolar Geometry based Learning of Multi-view Depth and Ego-Motion from\\ Monocular Sequences}

\begin{abstract}
Deep approaches to predict monocular depth and ego-motion have grown in recent years due to their ability to produce dense depth from monocular images. The main idea behind them is to optimize the photometric consistency over image sequences by warping one view into another, similar to direct visual odometry methods. One major drawback is that these methods infer depth from a single view, which might not effectively capture the relation between pixels. Moreover, simply minimizing the photometric loss does not ensure proper pixel correspondences, which is a key factor for accurate depth and pose estimations. \par

In contrast, we propose a 2-view depth network to infer the scene depth from consecutive frames, thereby learning inter-pixel relationships. To ensure better correspondences, thereby better geometric understanding, we propose incorporating epipolar constraints to make the learning more geometrically sound. We use the Essential matrix obtained using Nist\'er's Five Point Algorithm, to enforce meaningful geometric constraints, rather than using it as training labels. This allows us to use lesser no. of trainable parameters compared to state-of-the-art methods. The proposed method results in better depth images and pose estimates, which capture the scene structure and motion in a better way. Such a geometrically constrained learning performs successfully even in cases where simply minimizing the photometric error would fail.
\end{abstract}



\setcopyright{acmcopyright}

\acmDOI{10.1145/3293353.3293427}

\acmISBN{978-1-4503-6615-1/18/12}

\acmConference[ICVGIP 2018]{11th Indian Conference on Computer Vision, Graphics and Image Processing}{December 18--22 2018}{Hyderabad, India}
\acmYear{2018}
\copyrightyear{2018}

\acmPrice{15.00}

\acmBooktitle{11th Indian Conference on Computer Vision, Graphics and Image Processing (ICVGIP 2018), December 18--22, 2018, Hyderabad, India}

\maketitle

	\section{Introduction}
	\label{sec:intro}
 	In recent years, there has been an increasing trend in using deep networks for predicting dense depth and ego-motion from monocular sequences. Such methods, including the one proposed in this paper, make inferences of the scene by observing a lot of samples and inferring their understanding based on consistencies in the scene, similar to the way humans do. Building on top of the same idea of ensuring photometric consistency, some methods add additional supervised constraints in the form of ground-truth depth \cite{eigen2014depth,hoiem2005automatic,duggal2016hierarchical,saxena2007learning}, calibrated stereo rigs \cite{garg2016unsupervised,godard2017unsupervised} or ground-truth poses\cite{saxena2017exploring} or both pose and depth\cite{ummenhofer2017demon}. SE3-Nets\cite{byravan2017se3}  take it a step further and operate directly on pointcloud data to estimate rigid body motions.  \par
	
	SfMLearner\cite{zhou2017unsupervised} was one of the major developments in this field which popularized the ability of deep networks for the task at hand. They were the first method to do it in a purely monocular manner. In order to deal with non-rigid objects (cars, pedestrians etc.), they predict an "explainability mask" in order to discount regions that violate the static scene assumption. GeoNet\cite{yin2018geonet} and SfM-Net\cite{vijayanarasimhan2017sfm} tackle this issue by explicitly predicting object motions and incorporating optical. 
	
	However, one fallacy in these methods is that rather than taking a geometric approach to the problem, they play around with losses to get better performance. We build upon the SfMLearner pipeline and enrich it with the fundamentals of computer vision and multi-view geometry. We tackle the problem in a more geometrically meaningful way by constraining the correspondences to lie on their corresponding epipolar lines. This is done by weighting the losses using epipolar constraints with the Essential Matrix obtained from Nist\'er's Five Point Algorithm \cite{nister2004efficient}. This helps us account for violations of the static scene assumption, and also to tackle the problem of improper correspondence generation that arises by minimizing just the photometric loss. We make use of the Five Point Algorithm to help guide the training and improve the predictions. Moreover, rather than inferring depth from a single view, we try to learn inter-pixel relationships by predicting depth from two views in a sequence. Note that we do not use stereo pairs that have a wide baseline, but a sequence of monocular views.
	
	Our main contributions are twofold. The first is incorporating a 2-view depth prediction, rather than from a single view. One thing to note is that, by 2-view we mean two consecutive frames and not a pair of stereo images. Incorporating this helps improve the depth prediction, which is shown in Sec. \ref{ssec:pose_results}. Secondly, we incorporate epipolar constraints to make the learning more geometrically oriented. We do so by using the per-pixel epipolar distance as a weighting factor to help deal with occlusions and non-rigid objects. This not only enforces pixel level correspondences but also allows us to do away with the explainability mask thereby having lesser number of parameters to predict.\par

	\section{Background}
	\label{sec:bg}
	
	\subsection{Structure-from-Motion (SfM)}
	\label{ssec:sfm}
	\textbf{Structure-from-Motion} refers to the task of reconstructing an environment and recovering the camera motion from a sequence of images. Various methods exist that aim to tackle the problem as it is an age-old problem in vision\cite{wu2013towards,furukawa2010towards,sturm1996factorization,bhowmick2014divide} however, they usually require computation of accurate point correspondences for performing the task. A subset of this comes under the area of monocular SLAM, which involves solving the SfM problem in real-time. Visual odometry, is a smaller subset which doesn't involve structural estimation, but just camera motion estimation. These approaches could either be sparse\cite{mur2015orb,klein2007parallel,forster2014svo,maity2017edge,jose2015realtime}, semi-dense\cite{engel2014lsd,engel2018direct} or dense\cite{newcombe2011dtam,alismail2016enhancing}. The main issue that arises in these methods is that of improper correspondences in texture-less areas, or if there are occlusions or repeating patterns. In monocular approaches, performing it in a sparse manner is a well known topic but estimating dense monocular depth is much more complex. \par
	
	\subsection{Depth Image Based Warping}
	\label{ssec:dibr}
	In order to ensure that a reconstruction is accurate, one way is to ensure that the reprojected image of the predicted scene at a novel view point is consistent with what is observed at that point. This consistency is known as photometric consistency. One common approach for image warping in the context of deep networks using differentiable warping and bi-linear sampling \cite{jaderberg2015spatial} which have been in use for a variety of applications like learning optical flow \cite{jason2016back}, video prediction \cite{patraucean2015spatio} and image captioning \cite{johnson2016densecap}. We apply bi-linear sampling to calculate the warped image using scene depth and the relative transformation between two views. \par
	In this approach, given a warped pixel $\hat{p}$, its pixel value $I_s(\hat{p})$ is interpolated using 4 nearby pixel values of $\hat{p}$ (upper left and right, lower left and right) i.e. $\hat{I}_s(p) = I_s(\hat{p}) = \sum_{i} \sum_{j} w_{ij} I_s(p_{ij})$
	where $i\in\{\lfloor\hat{p}_x\rfloor,\lceil\hat{p}_x\rceil\}$, $j\in\{\lfloor\hat{p}_y\rfloor,\lceil\hat{p}_y\rceil\}$ and $w_{ij}$ is directly proportional to the proximity between $\hat{p}$ and $p_{ij}$ such that $\sum w_{ij} = 1$. Further explanation regarding this can be found in \cite{jaderberg2015spatial}.\par
	
	\subsection{Epipolar Geometry}
	\label{ssec:epipolar}
	We know that a pixel $p$ in an image corresponds to a ray in 3D, which is given by its normalized coordinates $\Tilde{p} = K^{-1}p$, where $K$ is the intrinsic calibration matrix of the camera. From a second view, the image of the first camera center is called the epipole and that of a ray is called the epipolar line. Given the corresponding pixel in the second view, it should lie on the corresponding epipolar line. This constraint on the pixels can be expressed using the Essential Matrix $E$ for calibrated cameras. The Essential Matrix contains information about the relative poses between the views. Detailed information regarding normalized coordinates, epipolar geometry, Essential Matrix etc, can be found in \cite{hartley2003multiple}. Given a pixel's normalized coordinates $\Tilde{p}$ and that of its corresponding pixel in a second view $\hat{\Tilde{p}}$, the relation between $\Tilde{p}$, $\hat{\Tilde{p}}$ and $E$ can be expressed as:
	
	\begin{equation}
	\label{eq:epipolar}
	\hat{\Tilde{p}}^TE\Tilde{p} = 0
	\end{equation}
	
	Here, $E\Tilde{p}$ is the epipolar line in the second view corresponding to a pixel $p$ in the first view. In most cases, there could be errors in capturing the pixel $p$, finding the corresponding pixel $\hat{p}$ or in estimating the Essential Matrix $E$. Therefore in most real world applications, rather than imposing Eq. \ref{eq:epipolar}, the value of $\hat{\Tilde{p}}^TE\Tilde{p}$ is minimized in a RANSAC\cite{fischler1981random} like setting. We refer to this value as the epipolar loss in the rest of the paper. \par 

	We refer to a pixel's homogeneous coordinates as $p$ and the normalized coordinates as $\Tilde{p}$ in the rest of the paper. We also refer to a corresponding pixel in a different view as $\hat{p}$ and in normalized coordinates as $\hat{\Tilde{p}}$.
	
	\subsection{Nist{\'e}r's 5 Point Algorithm}
	\label{ssec:5point}
	\textbf{Nist{\'e}r's 5 Point Algorithm} \cite{nister2004efficient} is currently the state-of-the-art solutions to the "Five Point Relative Pose" problem for estimating the Essential Matrix between two views. The problem can be stated as follows. Given the projections of five unknown points onto two unknown views, what are the possible configurations of points and cameras? The solution to this gives rise to the relative poses between the cameras and the points. Absolute scale, however, cannot be recovered from just images. Nist{\'e}r proposes a solution by solving a tenth degree polynomial in order to extract the Essential Matrix $E$ which can then be decomposed into the rotation $R$ and translation $t$ between the two views.

	\section{Proposed Approach}
	\label{sec:approach}
    
    \begin{figure*}[!h]
		\begin{center}
			\includegraphics[width=\textwidth]{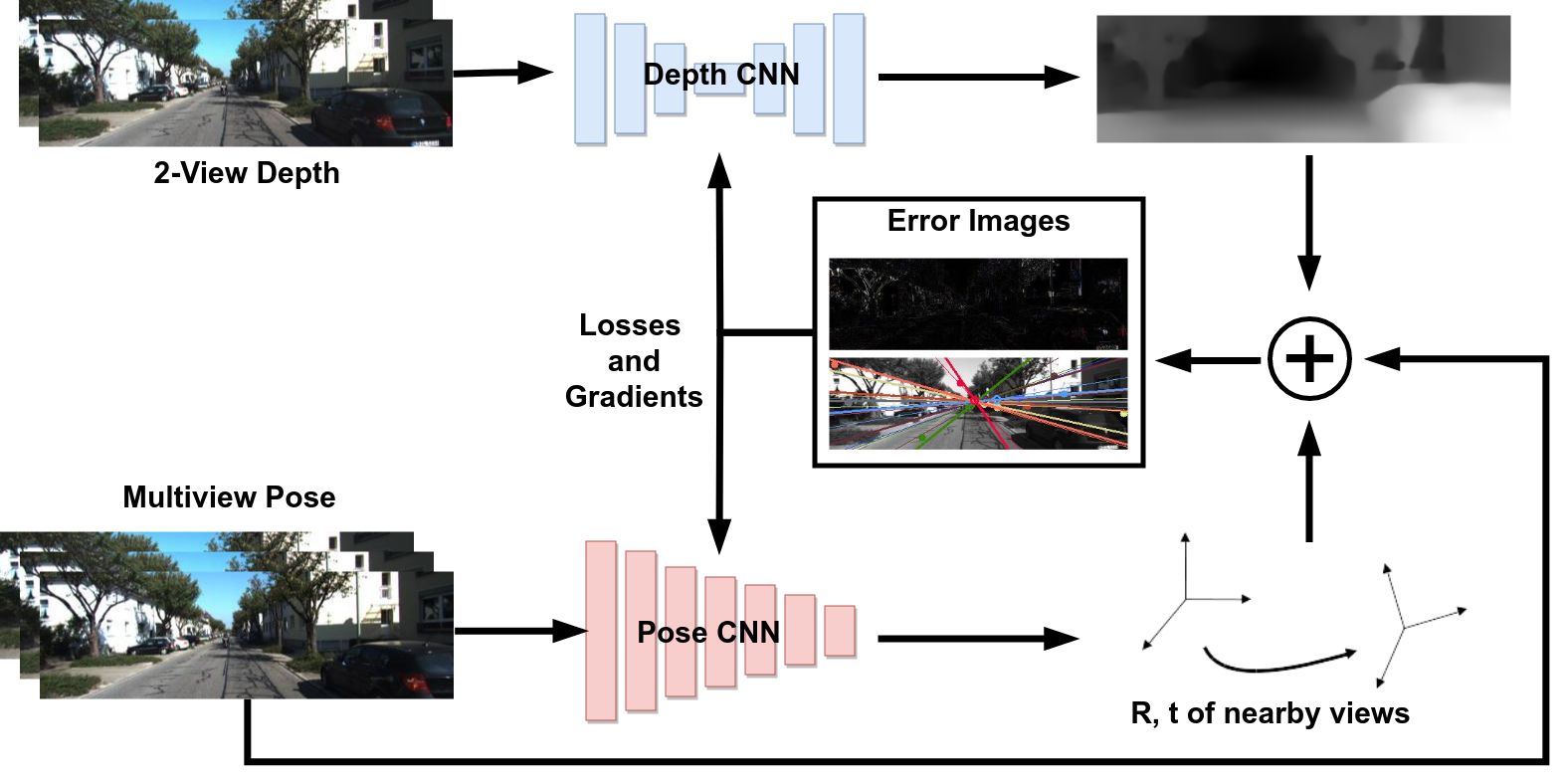}
			\caption{\small{Overview of the training procedure. The Depth CNN predicts the inverse depth for a target view by taking in the target view and a nearby image as the input. The Pose CNN predicts the relative poses of the source views from the target, which are then warped into the target frame using the relative poses and the scene depth and the photometric errors between multiple source-target frame pairs are minimized. These are weighted by the per-pixel epipolar loss.}}
			\label{fig:flow}
		\end{center}
	\end{figure*}
	
	We use 2 CNNs, one for inverse depth (which we refer to as depth network) and one for pose (pose network). Our Depth network takes a pair of consecutive images as input, rather than a single image, and calculates the depth of the scene as seen in the first image. The reason behind this is to leverage the relationship between pixels over multiple views to calculate the depth, rather than relying on learnt semantics of the scene in a single view. The Pose network takes an image sequence as input. The first image is the target view with respect to which the poses of the other images are calculated. Both networks are independent of each other and are trained jointly to effectively capture the coupling between scene depth and camera motion in learning based paradigm. \par
	The main idea behind the training, similar to that of previous works, is to ensure proper scene reconstruction between the source and target views based on the predicted depth and poses. We warp the source view into the target frame and minimize the photometric error between the synthesized image and the target image to ensure that the predicted depth and pose can recreate the scene effectively. Details about this warping process is given in Sec. \ref{ssec:warp} and \ref{ssec:vs}. Additionally, since we predict the depth using each source image as the second input image, we need to ensure consistency between the predicted depth images. This is explained in Sec. \ref{ssec:depth_cons}.\par
	
	For now, we consider mostly static scenes, i.e. where objects in the scene are rigid. SfMLearner predicts an "explainability mask" along with the pose, which 
	denotes the contribution of a pixel to the loss such that pixels of non-rigid objects have a low weight. 
	Instead, we use the epipolar loss to weight the pixels. This process is explained in Sec. \ref{ssec:epipolar_losses}. 
	
	\subsection{Image Warping}
	\label{ssec:warp}
	Given a pixel $p$ in normalized coordinates, and its depth $D(p)$, we transform it into the source frame using the relative pose and project it onto the source image's plane.
	\begin{equation}
	\label{eq:pixel_warping}
	\hat{p} = K(R_{t \rightarrow s}D(p)K^{-1}p + t_{t \rightarrow s})
	\end{equation}
	
	where $K$ is the camera calibration matrix, $D(p)$ is the depth of pixel $p$, $R_{t \rightarrow s}$ and $t_{t \rightarrow s}$ are the rotation and translation respectively from the target frame to the source frame. The homogeneous coordinates of $\hat{p}$ are continuous while we require integer values. Thus, we interpolate the values from nearby pixels, using bi-linear sampling, proposed by \cite{jaderberg2015spatial}, as explained in sec \ref{ssec:dibr}.\par

	\subsection{Novel View Synthesis}
	\label{ssec:vs}
	We use novel view synthesis using depth image based warping as the main supervisory signal. Given the per-pixel depth and the relative pose between images, we synthesize the image of the scene from a novel viewpoint. We minimize the photometric error between the warped image and the image at the given viewpoint. 
	
	Given a target view $I_t$ and $S$ source views $I_s$, we minimize the photometric error between the target view and the source view warped into the target's frame, denoted by $\hat{I}_s$. Mathematically, this can be described by Eq. \ref{eq:warp_loss}
	
	\begin{equation}
	\label{eq:warp_loss}
	L_{warp} = \frac{1}{N}\sum_{s=0}^S\sum_{p=0}^N|I_t(p) - \hat{I}_s(p)|
	\end{equation}
	where $N$ is the total number of pixels.

	\subsection{Depth Consistency}
	\label{ssec:depth_cons}
	Our depth network takes 2 views as input and outputs the depth w.r.t. the first image. In an iteration, we predict depths with each of the $S$ source views as the second input and use the respective outputs to warp a given source frame into the target frame. Since all the predicted depths are for the target image itself, they would need to be consistent with one another. Therefore we minimize the depth error between the predicted depth images obtained using all source images.
	
	\begin{equation}
	\label{eq:depth_loss}
	L_{depth} = \frac{1}{N}\sum_{i=0}^S\sum_{j=i+1}^S\sum_p|D_i(p) - D_j(p)|
	\end{equation}
	where $N$ is the total number of pixels.
	
	\subsection{Spatial Smoothing}
	\label{ssec:smooth}
	In order to tackle the issues of learning wrong depth values for texture-less regions, we try to ensure that the depth prediction is derived from spatially similar areas. One more thing to note is that depth discontinuities usually occur at object boundaries. We minimize $L1$ norm of the $2^{nd}$ order spatial gradients of the inverse depth of a pixel $\partial^2 d(p)$, weighted by the image laplacian at that pixel $\partial^2 I(p)$. This is to account for sudden changes in depth due to crossing of object boundaries and ensure a smooth change in the depth values. This is similar to what is done in \cite{godard2017unsupervised,wang2018learning}.
	\begin{equation}
	\label{eq:smooth_loss}
	L_{smooth} = \frac{1}{N}\sum_{p=0}^N\sum_{i\in\{x,y\}}\sum_{j\in\{x,y\}}|\partial_{ij} d(p)|e^{-|\partial_{ij} I(p)|}
	\end{equation}
	where $N$ is the total number of pixels.
	
	\subsection{Epipolar Constraints}
	\label{ssec:epipolar_losses}
	The problem with simply minimizing such photometric errors is that it doesn't take ambiguous pixels into consideration, such as those belonging to non-rigid objects, those which are occluded etc. Thus, we need to weight pixels appropriately based on whether they're properly projected or not. One way of ensuring correct projection is by checking if the corresponding pixel $\hat{p}$ satisfies epipolar constraints or not, according to Eq. \ref{eq:epipolar}.
		
	We impose epipolar constraints using the Essential Matrix obtained from Nist\'er's Five Point Algorithm \cite{nister2004efficient} using matches between features extracted using SiftGPU \cite{wu2007siftgpu}. This helps ensure that the warped pixels to lie on their corresponding epipolar line. This epipolar loss $\hat{\Tilde{p}}^TE\Tilde{p}$ is used to weight the above losses, where $E$ is the Essential Matrix obtained using the Five Point Algorithm. After weighting, the new photometric loss now becomes 
	
	\begin{equation}
	\label{eq:weighted_warp_loss}
	L_{warp} = \frac{1}{N}\sum_{s=0}^S\sum_{p=0}^N|I_t(p) - \hat{I}_s(p)|e^{|\hat{\Tilde{p}}^TE\Tilde{p}|}
	\end{equation}
	
	
	The reason behind this is that for a non-rigid object, even if the pixel is properly projected, the photometric error would be high. In order to ensure that such pixels are given a low weight, we weight them with their epipolar distance, which would be low if a pixel is properly projected. If the epipolar loss is high, it means that the projection is wrong, giving a high weight to the photometric loss, thereby increasing its overall penalty. This also helps in mitigating the problem of a pixel getting projected to a region of similar intensity by constraining it to lie along the epipolar line.\par 

	\subsection{Structural Similarity}
	
	Another well known and robust metric for measuring perceptual differences between two images is the Structural Similarity Index (SSIM) \cite{wang2004image}. It is widely applied in tasks that require comparing 2 images of the same scene, like comparing transmission quality. The photometric loss assumes brightness constancy which need not hold in all cases. Instead, SSIM considers three main factors, namely lunimance, constrast and structure, which provide a more robust measure for image similarity. Since SSIM needs to be maximized (with 1 as the maximum value), we minimize the below loss
	
	\begin{equation}
	\label{eq:ssim_loss}
	L_{ssim} = \sum_s\frac{1 - SSIM(I_t,\hat{I}_s)}{2}
	\end{equation}
	
	\subsection{Final Loss}
	Our final loss function is a weighted combination of the above loss functions summed over multiple image scales.
	\begin{equation}
	\label{eq:final_loss}
	\mathcal{L} = \sum_l (L_{warp}^l  + \lambda_{smooth}L_{smooth}^l + \lambda_{ssim}L_{ssim}^l +  \lambda_{depth}L_{depth}^l)
	\end{equation}
	
	where $l$ iterates over the different scale values and $\lambda_{smooth}$ $\lambda_{ssim}$  and $\lambda_{depth}$ are the the relative weights for the smoothness loss, SSIM loss and the depth consistency loss respectively. \par 
	Note that we don't minimize the epipolar loss but use it for weighting the other losses. This way the network tries to implicitly minimize it as it would lead to a reduction in the overall loss.


	\section{Implementation Details}
	\subsection{Neural Network Design}
	\label{ssec:networks}
	
	We use networks similar to that of SfMLearner, except we remove their "explainability mask" from the pose network. The network architectures that we use are shown in the appendix in Fig. \ref{fig:networks}.  
			\subsection{Depth Network}
			\label{ssec:depth_network}
			The design of the Depth CNN is similar to the one in \cite{zhou2017unsupervised}, which is inspired from DispNet\cite{mayer2016large}. It consists of an encoder-decoder network with skip connections from previous layers. The input is a pair of RGB images concatenated along the colour channel ($H \times W \times 6$) and the output is the inverse depth of the first image. The idea behind this is that rather than just learning semantic artifacts in the scene from a single view, the network can learn inter-pixel relationships and correspondences, similar to how optical flow networks are modelled. Even in Visual SLAM/Odometry methods, multiple images are used to predict depth, instead of a single image. \par
			
			Along with this, we also normalize the predicted inverse depth to have unit mean, to remove any scale ambiguity in the predicted depths. This is inspired from what is applied to the inverse depth of keyframes in LSD-SLAM\cite{engel2014lsd}.
			As mentioned in \cite{godard2018digging}, and as we observed in our experiments as well, performing a simple multi-scale estimation causes "holes" to develop in textureless regions. They argue that minimizing photometric error would allow the network to predict incorrect depth at a lower scale, which would still end up with a low photometric error at that scale due to the textureless property of the region but lead to a larger photometric error at a higher resolution. Thus they propose to overcome this by upsampling the depth images to the input's resolution and then calculating the errors. 
			
			\subsection{Pose Network}
			\label{ssec:pose_network}
			For the pose network, the target view and the source views are concatenated along the colour channel giving rise to an input layer of size $H \times W \times 3N$ where $N$ is the number of input views. The network predicts 6 DoF poses for each of the $N - 1$ source views relative to the target image. We modify the pose network proposed by \cite{zhou2017unsupervised} by removing their "explainability mask" thereby having to learn lesser parameters yet giving better performance. \par

		\begin{figure*}[h!]
			\centering
			\begin{tabular}{cccc}
				\textbf{Image}                               & \textbf{Ground truth}                      & \textbf{SfMLearner} & \textbf{Proposed Method} \\ 
				\includegraphics[width=0.23\textwidth]{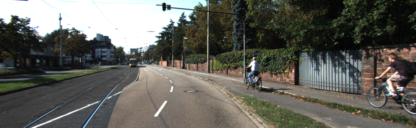} & \includegraphics[width=0.23\textwidth]{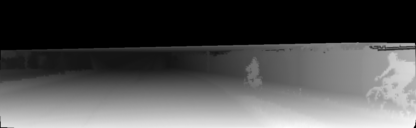} & \includegraphics[width=0.23\textwidth]{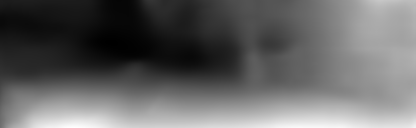} & \includegraphics[width=0.23\textwidth]{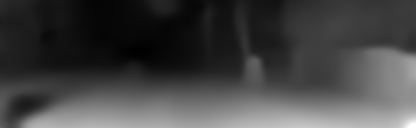} \\
				\includegraphics[width=0.23\textwidth]{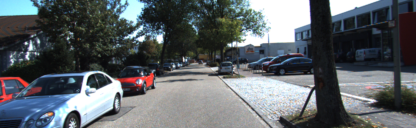} & \includegraphics[width=0.23\textwidth]{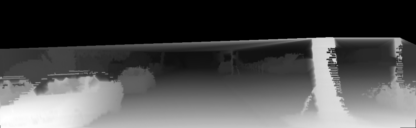} & \includegraphics[width=0.23\textwidth]{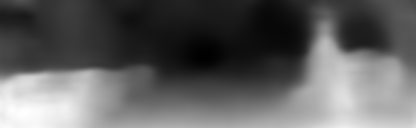} & \includegraphics[width=0.23\textwidth]{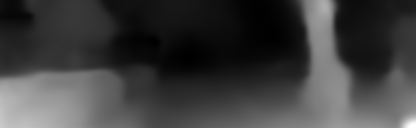} \\
				\includegraphics[width=0.23\textwidth]{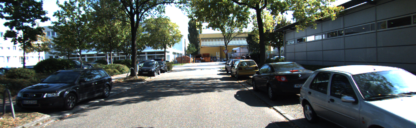} & \includegraphics[width=0.23\textwidth]{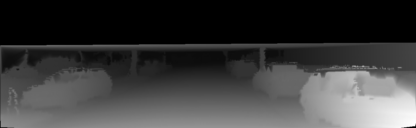} & \includegraphics[width=0.23\textwidth]{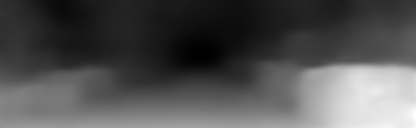} & \includegraphics[width=0.23\textwidth]{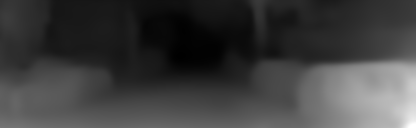} \\
				\includegraphics[width=0.23\textwidth]{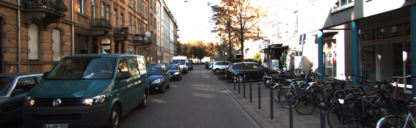} & \includegraphics[width=0.23\textwidth]{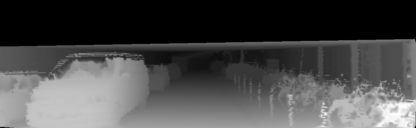} & \includegraphics[width=0.23\textwidth]{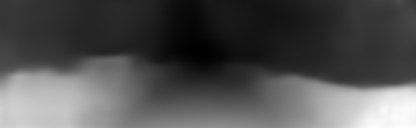} & \includegraphics[width=0.23\textwidth]{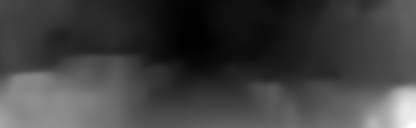} \\
				\includegraphics[width=0.23\textwidth]{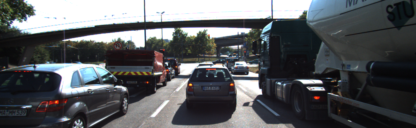} & \includegraphics[width=0.23\textwidth]{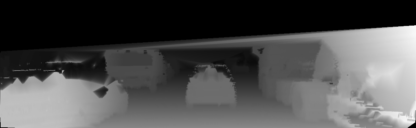} & \includegraphics[width=0.23\textwidth]{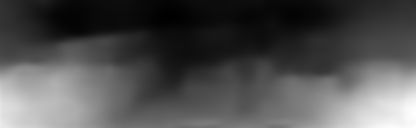} & \includegraphics[width=0.23\textwidth]{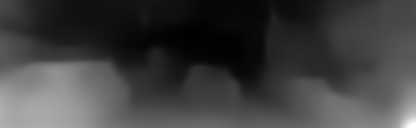} \\
				\includegraphics[width=0.23\textwidth]{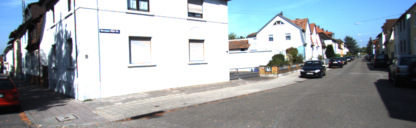} & \includegraphics[width=0.23\textwidth]{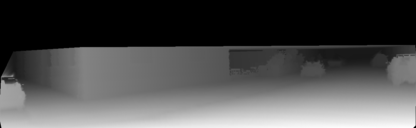} & \includegraphics[width=0.23\textwidth]{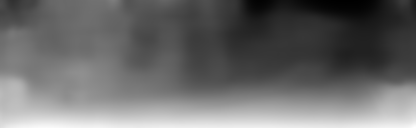} & \includegraphics[width=0.23\textwidth]{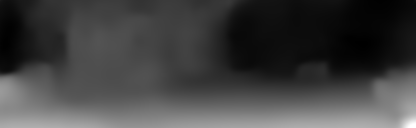} \\
				\includegraphics[width=0.23\textwidth]{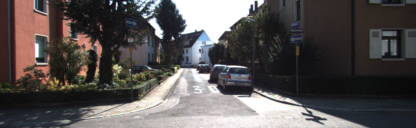} & \includegraphics[width=0.23\textwidth]{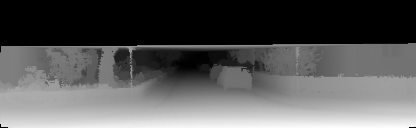} & \includegraphics[width=0.23\textwidth]{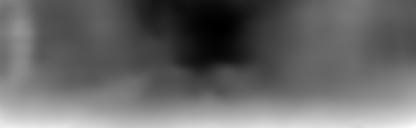} & \includegraphics[width=0.23\textwidth]{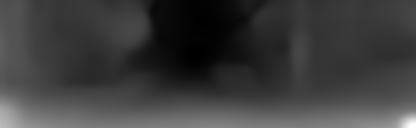} \\
				\includegraphics[width=0.23\textwidth]{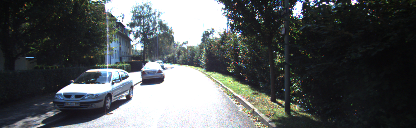} & \includegraphics[width=0.23\textwidth]{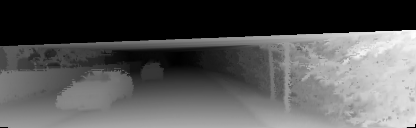} & \includegraphics[width=0.23\textwidth]{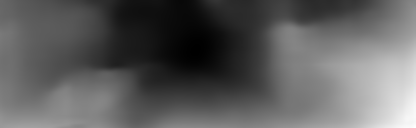} & \includegraphics[width=0.23\textwidth]{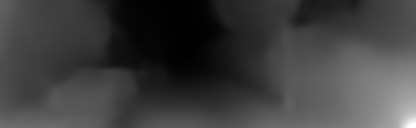} \\
    \arrayrulecolor{red}\hline\\
				\includegraphics[width=0.23\textwidth]{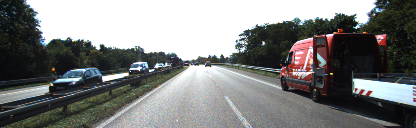} & \includegraphics[width=0.23\textwidth]{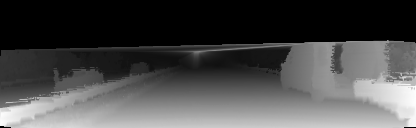} & \includegraphics[width=0.23\textwidth]{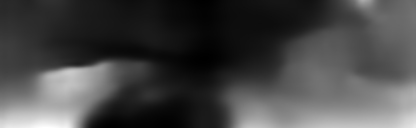} & \includegraphics[width=0.23\textwidth]{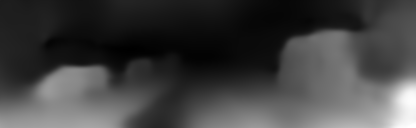} \\
				\includegraphics[width=0.23\textwidth]{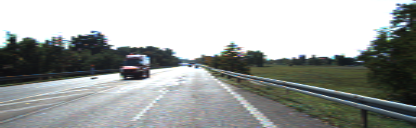} & \includegraphics[width=0.23\textwidth]{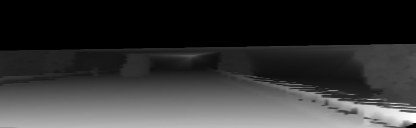} & \includegraphics[width=0.23\textwidth]{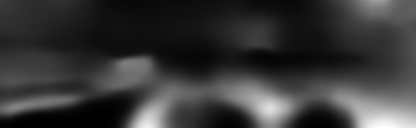} & \includegraphics[width=0.23\textwidth]{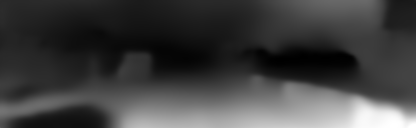} \\
				\includegraphics[width=0.23\textwidth]{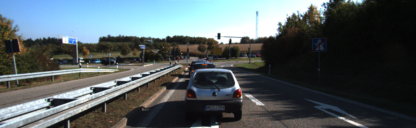} & \includegraphics[width=0.23\textwidth]{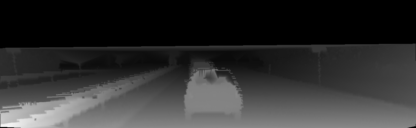} & \includegraphics[width=0.23\textwidth]{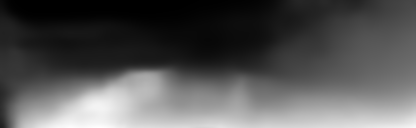} & \includegraphics[width=0.23\textwidth]{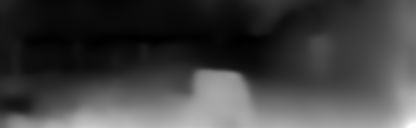} 
				
			\end{tabular}
			\caption{Results of depth estimation compared with SfMLearner. The ground truth is interpolated from sparse measurements for visualization purposes. Some of their main failure cases of SfMLearner are highlighted in the last 3 figures, such as large open spaces, texture-less regions, and when objects are present right in front of the camera. As it can be seen in the last 3 figures, our method performs better, providing more meaningful depth estimates even in such scenarios. (Pictures best viewed in color.)}
			\label{fig:depth_images}
		\end{figure*}	
	
	\subsection{Training}
	\label{ssec:training}
	We use Tensorflow \cite{abadi2016tensorflow} for implementing the system. We use batch normalization\cite{ioffe2015batch} for the non-output layers and make use of the Adam Optimizer \cite{kingma2014adam} with $\beta1 = 0.9$, $\beta2 = 0.999$ and a learning rate of 0.0002 and a mini-batch of size 4 for training our networks. We set the weights as $\lambda_{smooth} = 0.2$, $\lambda_{ssim} = 0.7$ and $\lambda_{depth} = 0.5$. The learning typically converges after 26 epochs.
	We use raw images from the KITTI dataset\cite{geiger2013vision}, with the split given by \cite{eigen2014depth}, having about 40K images totally. We exclude static scenes and  test image sequences from our training set leaving us with 33K images. We use 3 views as the input to the pose network with the middle image as our target image and the previous and next images as the source images. We use 2 views as input to the depth network with the middle image as the target view and predict it's depth with both the source images as the second input, one by one.\par
	
	

		\begin{table*}[h]
		\centering
		\resizebox{\textwidth}{!}{
		\begin{tabular}{|c|c|c|c|c|c|c|c|c|}
				\hline
				\multirow{2}{*}{\textbf{Method}}           & \multirow{2}{*}{\textbf{Supervision}} & \multicolumn{4}{c|}{\textbf{Error Metric} (lower is better)}     & \multicolumn{3}{c|}{\textbf{Accuracy Metric} (higher is better)} \\
				\cline{3-9}
				&                              & Abs. Rel. & Sq. Rel. & RMSE  & RMSE log &   $\delta < 1.25$         &$\delta < 1.25^2$            &     $\delta < 1.25^3$      \\
				\hline
				Train set mean                    & --                        & 0.403   & 5.53    & 8.709 & 0.403    & 0.593      & 0.776      & 0.878     \\
				Eigen et. al \cite{eigen2014depth} Coarse       & Depth                        & 0.214   & 1.605   & 6.563 & 0.292    & 0.673      & 0.884      & 0.957     \\
				Eigen et. al \cite{eigen2014depth} Fine         & Depth                        & 0.203   & 1.548   & 6.307 & 0.282    & 0.702      & 0.89       & 0.958     \\
				Liu et. al \cite{liu2016learning}               & Depth                        & 0.202   & 1.614   & 6.523 & 0.275    & 0.678      & 0.895      & 0.965     \\
				Godard et. al\cite{godard2017unsupervised}            & Stereo                        & 0.148   & 1.344   & 5.927 & 0.247    & 0.803      & 0.922      & 0.964     \\

				\hdashline
				Kuznietsov et. al\cite{kuznietsov2017semi} (Only Monocular)          & Mono                        & 0.308   & 9.367   & 8.700 & 0.367    & 0.752      & 0.904      & 0.952     \\
				Yang et. al\cite{yang2017unsupervised}           & Mono                        & 0.182   & 1.481   & 6.501 & 0.267    & 0.725      & 0.906      & 0.963     \\
				
				SfMLearner \cite{zhou2017unsupervised} (w/o explainability)   & Mono                             & 0.221   & 2.226   & 7.527 & 0.294    & 0.676      & 0.885      & 0.954     \\
				SfMLearner \cite{zhou2017unsupervised}                        & Mono                              & 0.208   & 1.768   & 6.856 & 0.283    & 0.678      & 0.885      & 0.957     \\
				SfMLearner \cite{zhou2017unsupervised} (updated from github)                        & Mono                              & 0.183   & 1.595   & 6.709 & 0.270    & 0.734      & 0.902      & 0.959     \\
				\hline
				\textbf{Ours}                        & Mono                              & \textbf{0.175}&	1.675&	\textbf{6.378}&	\textbf{0.255}&	\textbf{0.760}&	\textbf{0.916}&	\textbf{0.966}     \\
				\hline		
				Garg et. al \cite{garg2016unsupervised}      &       Stereo                       & 0.169   & 1.08    & 5.104 & 0.273    & 0.74       & 0.904      & 0.962     \\
\hdashline
				SfMLearner \cite{zhou2017unsupervised} (w/o explainability) &    Mono                          & 0.208   & 1.551   & 5.452 & 0.273    & 0.695      & 0.900        & 0.964     \\
				SfMLearner \cite{zhou2017unsupervised}                     &         Mono                     & 0.201   & 1.391   & 5.181 & 0.264    & 0.696      & 0.900        & 0.966    \\
				\hline
				\textbf{Ours}                      & Mono                              & \textbf{0.166}&	\textbf{1.213}&	\textbf{4.812}&	\textbf{0.239}&	\textbf{0.777}&	\textbf{0.928}&	\textbf{0.972}     \\
				\hline
			\end{tabular}
			}
			\vspace{0.01em}
			\caption{Single View Depth results using the split of \cite{eigen2014depth}. \cite{garg2016unsupervised} cap their depth at 50m which we show in the bottom part of the table. Further details about the error and accuracy metrics can be found in \cite{eigen2014depth}. The dashed line separates methods that use some form of supervision from purely monocular methods. Metrics are explained in the appendix in Sec. \ref{sec:depth_metrics}. Baseline numbers taken from \cite{zhou2017unsupervised,yang2017unsupervised,kuznietsov2017semi}.}.
			\label{table:depth_results}
		\end{table*}
\section{Results}
\label{sec:results}
	
	\subsection{Depth Estimation Results}
	\label{ssec:depth_results}
	We evaluate our performance on the 697 images provided by \cite{eigen2014depth}. We show our results in Table \ref{table:depth_results}. Our method's performance exceeds that of SfMLearner\cite{zhou2017unsupervised}, Yang et. al\cite{yang2017unsupervised}, Kuznietsov et. al\cite{kuznietsov2017semi} (only monocular) which are purely monocular. We also perform better than methods which use depth supervision \cite{eigen2014depth,liu2016learning}  and \cite{garg2016unsupervised} who use calibrated stereo supervision. We fall short of \cite{godard2017unsupervised}, who use calibrated stereo supervision along with left-right consistency, which makes their approach more robust.\par
	The images after the red line in Fig. \ref{fig:depth_images} are cases where our method performs better in places where SfMLearner fails, such as texture-less scenes and open regions. This shows the effectiveness of having 2-view depth prediction and using epipolar geometry to handle occlusions and non-rigidity. We provide sharper outputs as compared to SfMLearner, which can be seen in the  which is the result of using an edge-aware smoothness that helps capture the shape of objects in a better manner. We scale our depth predictions such that it matches the median of the ground truth. Further explanation about the metrics can be found is \cite{eigen2014depth}, which are given below in Sec. \ref{sec:depth_metrics}\par

		\begin{table*}
\centering
\begin{tabular}{|c|c|c|c|c|}
\hline
\multirow{3}{*}{\textbf{Seq}} & \multicolumn{2}{c|}{\textbf{Average Trajectory Error}} & \multicolumn{2}{c|}{\textbf{Average Translational Direction Error}} \\
\cline{2-5}
                                 & \multirow{2}{*}{\textbf{SfMLearner\cite{zhou2017unsupervised}}}      & \multirow{2}{*}{\textbf{Ours}}              & \textbf{Five Point Algorithm\cite{nister2004efficient}}           & \multirow{2}{*}{\textbf{Ours}}\\             
                                 &&&\textbf{using SiftGPU\cite{wu2007siftgpu}}&\\
\hline
\textbf{00}                  & 0.5099$\pm$0.2471            & \textbf{0.4967$\pm$0.1787}     & 0.0084$\pm$0.0821                 & \textbf{0.0040$\pm$0.0155}           \\
\textbf{01}                  & 1.2290$\pm$0.2518            & \textbf{1.1458$\pm$0.2175}     & 0.0061$\pm$0.0807                 & \textbf{0.0033$\pm$0.0077}           \\
\textbf{02}                  & 0.6330$\pm$0.2328            & 0.6512$\pm$0.1806              & 0.0035$\pm$0.0509                 & \textbf{0.0021$\pm$0.0026}           \\
\textbf{03}                  & 0.3767$\pm$0.1527            & \textbf{0.3583$\pm$0.1254}     & 0.0142$\pm$0.1611                 & \textbf{0.0027$\pm$0.0042}           \\
\textbf{04}                  & 0.4869$\pm$0.0537            & 0.6404$\pm$0.0607              & 0.0182$\pm$0.2131                 & \textbf{0.0002$\pm$0.0007}           \\
\textbf{05}                  & 0.5013$\pm$0.2564            & \textbf{0.4930$\pm$0.1974}     & 0.0130$\pm$0.0945                 & \textbf{0.0044$\pm$0.0044}           \\
\textbf{06}                  & 0.5027$\pm$0.2605            & 0.5384$\pm$0.1627              & 0.0130$\pm$0.1591                 & \textbf{0.0080$\pm$0.0688}           \\
\textbf{07}                  & 0.4337$\pm$0.3254            & \textbf{0.4032$\pm$0.2380}     & 0.0508$\pm$0.2453                 & \textbf{0.0114$\pm$0.0430}           \\
\textbf{08}                  & 0.4824$\pm$0.2396            & \textbf{0.4708$\pm$0.1827}     & 0.0091$\pm$0.0646                 & \textbf{0.0037$\pm$0.0058}           \\
\textbf{09}                  & 0.6652$\pm$0.2863            & \textbf{0.6280$\pm$0.2028}     & 0.0204$\pm$0.1722                 & \textbf{0.0073$\pm$0.0211}           \\
\textbf{10}                  & 0.4672$\pm$0.2398            & \textbf{0.4185$\pm$0.1791}              & 0.0200$\pm$0.1241                 & \textbf{0.0040$\pm$0.0105}   \\
\hline
\end{tabular}
\vspace{0.3em}
\caption{Average Trajectory Error (ATE) compared with SfMLearner and Average Translational Direction Error (ATDE) compared with the Five Point Algorithm averaged over 3 frame snippets on the KITTI Visual Odometry Dataset \cite{geiger2012CVPR}. The ATE is shown in meters and the ATDE, in radians. All values are reported as mean $\pm$ std. dev.}
\label{table:pose_error}
\end{table*}			

\subsection{Depth Evaluation Metrics}
	\label{sec:depth_metrics}
	Given the predicted depth $\hat{y}_i$ and the corresponding ground truth depth $y_i^*$ for the $i^{th}$ image, we use the following error metrics and accuracy metrics.
		\subsubsection{Error Metrics}
		\label{ssec:error_metrics}
		\begin{itemize}
		\item Absolute Relative Difference (Abs. Rel.): $\frac{1}{N}\sum_{i = 1}^N|\hat{y}_i - y_i^*|/y_i^*$
		\item Squared Relative Difference (Sq. Rel.): $\frac{1}{N}\sum_{i = 1}^N||\hat{y}_i - y_i^*||^2/y_i^*$
		\item Root Mean Squared Error (RMSE): $\sqrt{\frac{1}{N}\sum_{i = 1}^N||\hat{y}_i - y_i^{*}||^2}$
		\item Logarithmic RMSE (RMSE log): $\sqrt{\frac{1}{N}\sum_{i = 1}^N||\log \hat{y}_i - \log y_i^*||^2}$
		\end{itemize}
		
		\subsubsection{Accuracy Metrics}
		\label{ssec:acc_metrics}
		We calculate the accuracy metric as the percentage of images for which the value of $\delta$, defined as $\delta = \max(\frac{\hat{y}_i}{y_i^*}, \frac{y_i^*}{\hat{y}_i})$, is lesser than a threshold $th$. In our case, we choose three values for the threshold $th$ which are $1.25$, $1.25^2$ and $1.25^3$.

		
		
				\subsection{Pose Estimation Results}
		\label{ssec:pose_results}
		
		For our pose estimation experiments, we use the KITTI Visual Odometry Benchmark dataset \cite{geiger2012CVPR}. Only 11 sequences (00-10) have the associated ground truth data, on which we show our results. We use a sequence of 3 views as the input to the pose network with the middle view as the target view. Each image is of size $1271\times376$ which we scale down to a size of $416\times128$ for both pose estimation and depth estimation experiments.\par
		
		We show the Average Trajectory Error (ATE) and Average Translational Direction Error (ATDE) averaged over 3 frame intervals. Before comparison, the scale is first corrected to best align it with the ground truth after which the ATE is computed. Since the ATDE is only comparing the angle between the directions of translations, we do not correct the scale of the poses. \par 
		
		Table \ref{table:pose_error} shows the results of our pose estimation. We perform better than SfMLearner\footnote{using the model provided at \url{github.com/tinghuiz/SfMLearner}} on an average in terms of the ATE showing that adding meaningful geometric constraints helps get better estimates as compared to minimzing just the reprojection error. We perform better on all runs compared to the relative poses obtained from the Five Point Algorithm in terms of the ATDE. \par 
		This rises from the fact that, we have additional constraints of depth and image warping that help give a better estimation of the direction of motion, whereas the Five point Algorithm uses only sparse point correspondences between the images. Moreover, the Five point Algorithm itself is slightly erroneous due to inaccuracies arising in the feature matching or in the RANSAC based estimation of the essential matrix. Despite being given slightly erroneous estimates of the essential matrix, incorporating image reconstruction as the main goal helps in overcoming erroneous predictions. 

\begin{table*}[h]
		\centering
		\resizebox{\textwidth}{!}{
		\begin{tabular}{|c|c|c|c|c|c|c|c|}
				\hline
				\multirow{2}{*}{\textbf{Method}}           & \multicolumn{4}{c|}{\textbf{Error Metric} (lower is better)}     & \multicolumn{3}{c|}{\textbf{Accuracy Metric} (higher is better)} \\
				\cline{2-8}
				                              & Abs. Rel. & Sq. Rel. & RMSE  & RMSE log &   $\delta < 1.25$         &$\delta < 1.25^2$            &     $\delta < 1.25^3$      \\
				\hline
				SfMLearner (w/o explainability)                                & 0.221   & 2.226   & 7.527 & 0.294    & 0.676      & 0.885      & 0.954     \\
				Ours (no-epi)                                                      &0.199 &1.548	&6.314	&0.274	&0.697	&0.901	&0.964	     \\
				Ours (single-view)                                                      &0.181 &1.520	&6.080	&0.260	&0.747	&0.914	&0.965	     \\
				Ours ($1^{st}$ order)                                                      &0.190 &1.440	&6.144	&0.269	&0.714	&0.906	&0.965	     \\
				Ours (final)                                                      &0.175 &1.675	&6.378	&0.255	&0.761	&0.916	&0.966	     \\
				\hline
			\end{tabular}
			}
			\vspace{0.01em}
			\caption{Ablative study on the effect of different losses in depth estimation. We show our results using the split of \cite{eigen2014depth} while removing the proposed losses. We compare our method with  SfMLearner (w/o explainability) which is essentially similar to stripping our method of the proposed losses and using a different smoothness loss.}
			\label{table:ablation_results}
		\end{table*}	
\section{Ablation Study}
In order to see what is the contribution of our proposed loss to the learning process, we perform an ablation study on the depth estimation by considering variants of our proposed approach and training the networks using the Eigen\cite{eigen2014depth} split. We show the results of studying the effects of epipolar constraints, 2-view depth prediction and the second order edge-based smoothness. \par

In order to see the effectiveness of our method, we first replace the second order edge-aware smoothness with a first order edge-based smoothness. We call this as "Ours ($1^{st}$ order)". Then we remove the epipolar loss which is similar to having an SfMLearner pipeline with 2 views as input to the depth network. We denote this as "Ours (no-epi)". Then in order to see the contribution of our proposed idea of using 2 views as input to the depth network, we train a variant with a single view depth having epipolar constraints and a first order edge-based smoothness. This is denoted as "Ours (single-view)".\par 

Any further removal would just result in having SfMLearner without the explainability mask, which we consider as our baseline. This is essentially equivalent to removing our proposed improvements and replacing the edge-aware depth with a simpler second order smoothness loss. Finally, we further investigate the possibility of using 3 views as the input to the depth network as well.\par

\begin{figure*}[h!]
			\centering
			\subfigure[]{
        	\includegraphics[width=0.3\textwidth]{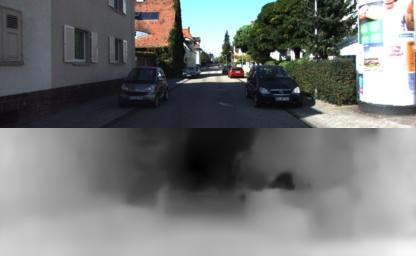}
        	\label{subfig:3vdepth_normal_1}
        	}
            \subfigure[]{
        	\includegraphics[width=0.3\textwidth]{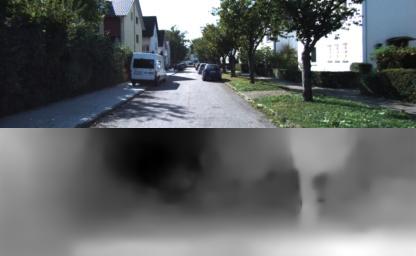}
        	\label{subfig:3vdepth_normal_2}
        	} 
        	\subfigure[]{
        	\includegraphics[width=0.3\textwidth]{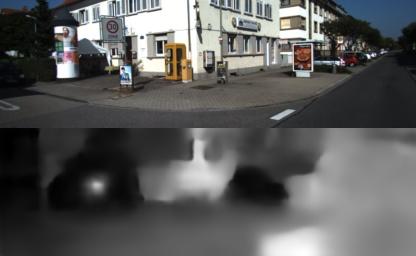}
        	\label{subfig:3vdepth_normal_turn}
        	}
			
			\caption{Results of 3-view depth estimation. As it can be seen, though the coarse structure of the scene is detected, objects are still not as finely reconstructed. The third image is an example where the car is taking a turn, which gives haphazard depth outputs.}
			\label{fig:3vdepth_normal}
\end{figure*}

\begin{figure*}[h!]
			\centering
			\subfigure[]{
        	\includegraphics[width=0.3\textwidth]{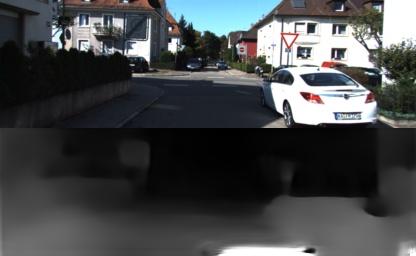}
        	\label{subfig:3vdepth_large_1}
        	}
            \subfigure[]{
        	\includegraphics[width=0.3\textwidth]{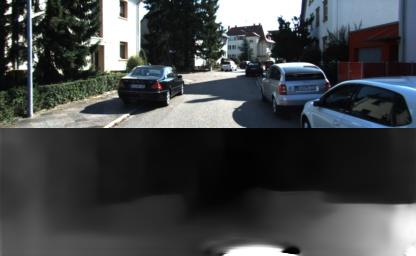}
        	\label{subfig:3vdepth_large_2}
        	} 
        	\subfigure[]{
        	\includegraphics[width=0.3\textwidth]{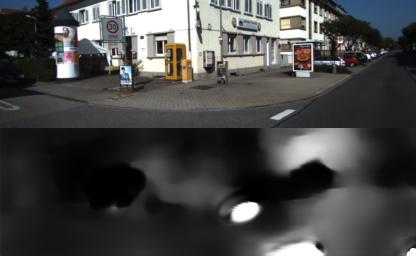}
        	\label{subfig:3vdepth_large_turn}
        	}
			
			\caption{Results of 3-view depth estimation with using larger convolutional filter sizes. It gives a smoother depth output as before, however, it still isn't able to fully capture the scene depth. The third image is an example where the car is taking a turn, which still gives haphazard depth outputs.}
			\label{fig:3vdepth_large}
\end{figure*}

The results of the study are shown in Table \ref{table:ablation_results}. All our variants perform significantly better than the standard SfMLearner without the explainability mask showing that our method has a positive effect on the learning. Just adding 2-view depth prediction without epipolar constraints (Ours no-epi) gives a significant improvement showing that using multiple views provide better depth estimates, compared to a single view. Incorporating just the epipolar loss (Ours single-view) improves the output drastically, showing that geometrically meaningful constraints provide better depth outputs. \par.

When we combine the two, it doesn't lead to a drastic improvement over the individual methods, however providing a second order smoothness improves the results as compared to a first order smoothness. A first order smoothness implies having a constant depth change, which isn't necessarily true. For parts of the image that are closer to the camera, the depth varies slowly, whereas for those which are further away, the depth variations are larger. Therefore, having a second order smoothness captures this change in depth variation rather than just the change in depth. \par

\subsection{Three-view Depth Prediction}
One more variant which we tried out was using 3 views for the depth prediction as well. Though it was giving visually understandable results, our 2-view variant was giving better depth estimates. One observation is that during turning, the depth outputs would deteriorate at a much larger scale. A possible explanation for this could be that compared to the motion between 2 views, while using 3 views, there is a larger amount of motion, and thereby lesser amount of overlap between images. Though the motion is not too large, it is large enough to escape the field of view of the convolutional filter in an input, since the views are stacked together and given to the network. Some sample outputs of using a 3 view depth prediction are shown in Fig. \ref{fig:3vdepth_normal}.

In order to test our hypothesis regarding the filter sizes, we tried increasing the filter sizes and performing the depth prediction. More specifically, we increase each of the filter sizes by a value of 4. By doing so, we effectively increase the perceptive field of view of a filter thus allowing it to accumulate information from a larger area of pixels. This way, it would be able to properly "see" a pixel across multiple views, which would otherwise fall outside the field of view of a filter. We observed that this leads to smoother depth images as compared to using smaller filter sizes. However, it still didn't perform as well as the 2-view variant, and it ended up developing a few unwanted holes as well. These results are shown in Fig. \ref{fig:3vdepth_large}.

		\section{Conclusion and Future Work}
		\label{sec:conc}
		We build upon a previous unsupervised method for learning deep monocular visual odometry and depth by leveraging the fact that depth estimation could be made more robust by using multiple views rather than a single view. Along with this, we incorporate epipolar constraints to help make the learning  more geometrically meaningful while using lesser number of trainable parameters. Our method is able to predict depth with higher accuracy along with giving sharper depth estimates and better pose estimates. Although increasing the number of inputs for depth prediction gave a good output in 2 views, it's 3-view counterpart wasn't able to perform as well. This would be an interesting problem to look into for improving it, either by architectural changes in the depth network or by incorporating a post-processing optimization on top of the networks.  
		
		The current method however only performs pixel level inferences. A higher scene level understanding can be obtained by integrating semantics of the scene to get better correlation between objects in the scene and the depth and ego-motion estimates. This is similar to using semantic motion segmentation \cite{haque2017joint,haque2017temporal}. Architectural changes could also be leveraged to get a stronger coupling between depth and pose by having a single network predicting both pose and depth in order to allow the network to be able to learn representations that capture the complex relation between both camera motion and scene depth. 
		


\bibliographystyle{ACM-Reference-Format}
\bibliography{references.bib}
\clearpage
\chapter{ \centering \textbf{\noindent\resizebox{\textwidth}{!}{Appendix for "Epipolar Geometry based Learning of Multi-view Depth}}\\
\hspace{0.175\textwidth}\textbf{\noindent\resizebox{0.65\textwidth}{!}{and Ego-Motion from Monocular Sequences"}}}
\appendix
\setcounter{figure}{0}
\renewcommand\thefigure{A\arabic{figure}}

\begin{appendix}
\input{nets.tex}
\end{appendix}

\end{document}